%% file: main.tex
\documentclass[runningheads]{llncs}

\usepackage{eccv}

\usepackage{eccvabbrv}

\usepackage{graphicx}
\usepackage{booktabs}

\usepackage[accsupp]{axessibility}  

\usepackage[pagebackref,breaklinks,colorlinks,citecolor=eccvblue]{hyperref}

\usepackage{orcidlink}

\usepackage{bbm}
\usepackage{algorithm}
\usepackage{algpseudocode}
\usepackage{amsmath}
\usepackage{amsfonts}
\usepackage{tabularx}
\usepackage{booktabs}
\usepackage{adjustbox}
\usepackage{wasysym}
\usepackage{enumitem}

\newcolumntype{L}[1]{>{\raggedright\arraybackslash}p{#1}}
\newcolumntype{C}[1]{>{\centering\arraybackslash}p{#1}}
\newcolumntype{R}[1]{>{\raggedleft\arraybackslash}p{#1}}
\usepackage{wrapfig}
\begin{document}

\title{R.A.C.E.~: Robust Adversarial Concept Erasure for Secure Text-to-Image Diffusion Model} 

\titlerunning{R.A.C.E. for Secure T2I Diffusion}

\author{Changhoon Kim\thanks{These authors contributed equally to this work.}\inst{1}\orcidlink{0009-0000-5850-6483} \quad
Kyle Min$^{\star}$\inst{2}\orcidlink{0000-0002-3978-918X} \quad Yezhou Yang\inst{1}\orcidlink{0000-0003-0126-8976}
}

\authorrunning{C. Kim$^{\star}$, K. Min$^{\star}$, Y. Yang}

\institute{Arizona State University \and Intel Labs \\
\email{\{kch,yz.yang\}@asu.edu} \quad \email{kyle.min@intel.com}
}

\maketitle

\begin{abstract}
In the evolving landscape of text-to-image (T2I) diffusion models, the remarkable capability to generate high-quality images from textual descriptions faces challenges with the potential misuse of reproducing sensitive content. To address this critical issue, we introduce \textbf{R}obust \textbf{A}dversarial \textbf{C}oncept \textbf{E}rase (RACE), a novel approach designed to mitigate these risks by enhancing the robustness of concept erasure method for T2I models. RACE utilizes a sophisticated adversarial training framework to identify and mitigate adversarial text embeddings, significantly reducing the Attack Success Rate (ASR). Impressively, RACE achieves a 30 percentage point reduction in ASR for the ``nudity'' concept against the leading white-box attack method. Our extensive evaluations demonstrate RACE's effectiveness in defending against both white-box and black-box attacks, marking a significant advancement in protecting T2I diffusion models from generating inappropriate or misleading imagery. 
This work underlines the essential need for proactive defense measures in adapting to the rapidly advancing field of adversarial challenges. Our code is publicly available: \url{https://github.com/chkimmmmm/R.A.C.E.}

\end{abstract}

\section{Introduction} \label{sec:intro}

\begin{figure}[t!]
\centering
  \includegraphics[width=0.75\linewidth]{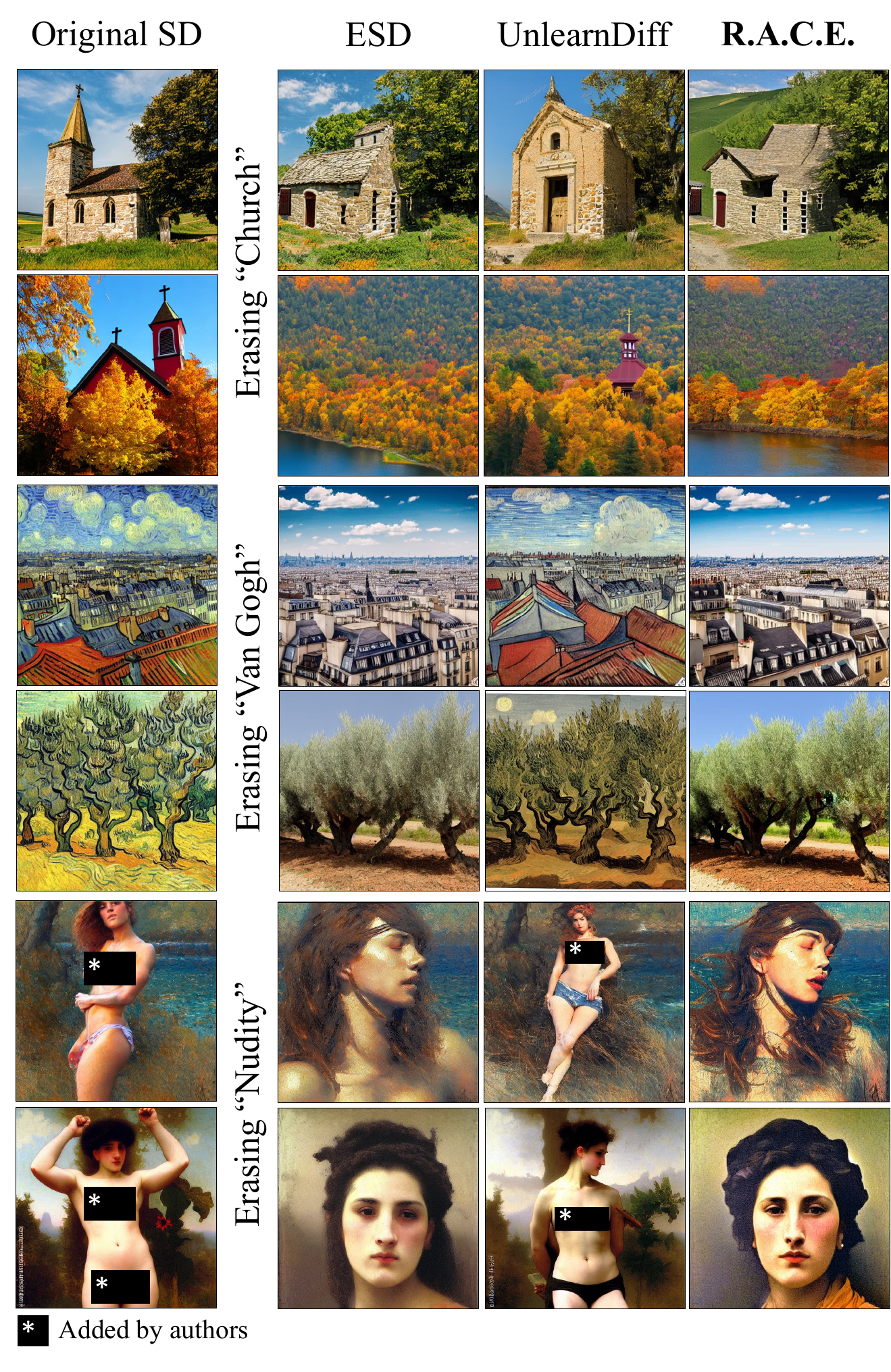}
  \caption{
  Comparative demonstration of concept erasure, red teaming, and robust erasure within T2I diffusion models. The ESD method~\cite{ESD} removes targeted concepts from the original SD outputs, yet these concepts can be reconstructed using UnlearnDiff~\cite{to_generate_or_not}. Our proposed \textbf{R.A.C.E.} method showcases enhanced robustness against such red teaming reconstruction efforts.
  }
  \label{fig:Fig1}
  \vspace{-0.5cm}
\end{figure}

The field of text-to-image (T2I) diffusion models has garnered significant attention for their ability to produce high-quality images that can be adaptively generated from textual descriptions~\cite{stable_diffusion,ramesh2022hierarchical}. This advancement is predicated on the training of T2I models with extensive datasets, often encompassing a range of content including copyrighted, explicit, and private materials~\cite{schuhmann2022laion, Somepalli2022DiffusionAO, Somepalli2023UnderstandingAM}. Consequently, these models possess the capacity to inadvertently replicate protected images, potentially without user awareness~\cite{Somepalli2022DiffusionAO, Somepalli2023UnderstandingAM, Carlini2023ExtractingTD}. The misuse of T2I models by malicious actors for misinformation or public opinion manipulation presents a significant concern~\cite{devlin2023faketrump,marcelo2023pentagon}.

In response to the challenges posed by the malicious exploitation of generative models, Stable Diffusion (SD)\cite{stable_diffusion} has integrated a safety checker\cite{rando2022red} and advocates for the utilization of a watermarking module~\cite{rivagan}. Despite these initiatives, the reliance on post-hoc interventions presents limitations due to their potential for circumvention~\cite{WOUAF,stable_signature}. Consequently, the research community is pivoting towards formulating methodologies that embed safety protocols directly within the image generation pipeline. This approach ensures that content security is an intrinsic aspect of image creation~\cite{tree-ring, yu2020responsible, yu2021artificial, kim2020decentralized, nie2023attributing,WOUAF,stable_signature}. 
Although these methods are valuable for identifying the source of content after an incident, their reactive nature highlights the necessity for proactive strategies. This includes pioneering techniques attempting early removal of sensitive content from T2I models~\cite{Ablating_Concept,ESD}, thereby preempting the production of inappropriate or harmful material.

To address the challenge of removing undesirable content from T2I models, even when users attempt to circumvent content restrictions, recent research has focused on the development of concept erasure techniques within T2I diffusion models~\cite{ESD,Ablating_Concept,Zhang2023ForgetMeNotLT,Heng2023SelectiveAA}. These techniques primarily aim to eliminate specific concepts (e.g., ``nudity'') by altering the text embeddings associated with these concepts to neutral representations. Despite these efforts, there remains a vulnerability wherein erased concepts can be reconstructed. This is achieved by identifying text tokens that closely align with the visual embeddings of the targeted concepts, thus enabling the regeneration of prohibited content~\cite{p4d,to_generate_or_not,Tsai2023RingABellHR,hard_prompt,Yang2023MMADiffusionMA}. 
This issue is illustrated in Fig.~\ref{fig:Fig1}, demonstrating that even with concept erasure, T2I models can be manipulated through prompt modification to regenerate the restricted content.
This underscores the imperative for a more robust concept erasure methodology that can withstand such reconstruction attempts, ensuring the integrity of content generation within T2I models.

Acknowledging the imperative for enhanced concept erasure methodologies within T2I models, we pose a critical question: \textit{``Is it feasible to develop a concept erasure approach that is resilient against reconstruction efforts?''} In pursuit of this, we introduce \textbf{R.A.C.E.} (\textbf{R}obust \textbf{A}dversarial \textbf{C}oncept \textbf{E}rase), a novel strategy aimed at bolstering the resilience of concept erasure techniques against adversarial manipulations, as delineated in Algorithm~\ref{alg:our_method}. At the heart of RACE lies an adversarial training framework, leveraging insights from the effort of adversarial robustness~\cite{PGD}. Our method effectively identifies adversarial text embeddings capable of reconstructing erased concepts and then facilitates their integration into the T2I concept erasure workflow.

A pivotal aspect of RACE is its ability to efficiently uncover adversarial text embeddings within a single time step of the diffusion process, an approach elaborated in Section~\ref{subsec:adv_training}. This efficiency not only streamlines the process of identifying adversarial examples but also facilitates the integration of our adversarial attack mechanism into the concept erasure workflow. To demonstrate the robustness of RACE, we have carried out an extensive array of experiments. These experiments validate the effectiveness of RACE in countering diverse red teaming strategies, with detailed results presented in Section~\ref{sec:experiments}. Our empirical investigations underscore the capacity of RACE to significantly enhance the robustness of T2I models against both white-box and black-box attacks across a broad spectrum of target concepts, including artistic, explicit, and object categories.

We summarize the three main contributions here:
\begin{itemize}[noitemsep,nolistsep]
    \item We are the first to present, to our knowledge, an adversarial training approach specifically designed to fortify concept erasure methods against prompt-based adversarial attacks without introducing additional modules.
    \item Our method, RACE, implements a computationally efficient adversarial attack method that can be plugged into the concept erasing method.  
    \item We show RACE significantly improves T2I models' robustness against prompts based on white/black box attacks.    
\end{itemize}


\section{Related Works}
For a comprehensive overview of additional related works, please refer to the supplementary material.\\

\noindent
\textbf{Text-to-Image Synthesis.} The field of generative models has seen remarkable advancements, notably extending their capabilities beyond generating photorealistic images~\cite{karras2020analyzing,esser2021taming} to include Text-to-Image (T2I) synthesis~\cite{nichol2022glide,Patel2023ECLIPSEAR,saharia2022photorealistic,ramesh2022hierarchical, stable_diffusion}. This progress has led to the development of fine-tuning techniques that allow for the customization of T2I models to user-specific needs~\cite{Gal2022AnII, Ruiz2022DreamBoothFT,Kumari2022MultiConceptCO,Patel2024ECLIPSEMP,Wei2023ELITEEV}, thereby enabling the creation of highly realistic images that align closely with textual prompts. However, the potential for misuse by malicious entities, using these models for purposes such as spreading misinformation~\cite{devlin2023faketrump,marcelo2023pentagon}, raises significant concerns. This underscores the urgency of devising protective measures to mitigate the risk of such exploitations.
\\

\noindent
\textbf{Advanced Techniques in Concept Erasure for T2I Diffusion Models.}
Within the realm of machine unlearning, concept erasure for T2I Diffusion models has recently emerged as a critical area of research, focusing on the removal of sensitive or copyrighted concepts from T2I models. Methods to achieve this include guiding the image generation process or adjusting the model's weights to exclude these elements~\cite{Schramowski2022SafeLD, Li2024GetWY, Ni2023ORESOR, ESD,Ablating_Concept,Zhang2023ForgetMeNotLT,Heng2023SelectiveAA}. Notably, techniques by Gandikota et al.\cite{ESD} and Kumari et al.\cite{Ablating_Concept} involve mapping sensitive concepts to null entities or benign equivalents by fine-tuning the weights of Stable Diffusion (SD)~\cite{stable_diffusion} models, effectively preventing the generation of undesirable content. Despite these advancements, red teaming methods have exposed potential loopholes, indicating that erased concepts might be regenerated through meticulously designed text prompts. Addressing this issue, our work contributes an adversarial training strategy aimed at bolstering the resilience of Stable Diffusion models against such text-prompt-based attacks~\cite{p4d,to_generate_or_not,hard_prompt}, thereby enhancing the security and integrity of the content generation.\\

\noindent
\textbf{Robustness Evaluation via Red Teaming in T2I Models.}
While various safety measures have been proposed to shield SD models from misuse, red teaming strategies reveal vulnerabilities that still allow for circumvention. Research has demonstrated that techniques like Textual Inversion~\cite{Gal2022AnII} can be exploited to regenerate content previously erased from SD models~\cite{pham2024circumventing}, prompting the development of countermeasures aimed at safeguarding against such inversions~\cite{Wu2023BackdooringTI,Zheng2023IMMAIT}. In real-world applications, T2I services such as Midjourney predominantly rely on user-provided text prompts, making them susceptible to prompt-based red teaming attacks~\cite{p4d,to_generate_or_not,Tsai2023RingABellHR,hard_prompt,Yang2023MMADiffusionMA}. These methods employ sophisticated prompt optimization techniques to restore images containing erased content, with their efficacy contingent upon the level of model access—categorized into white-box approaches, which utilize SD's U-Net~\cite{ronneberger2015u} for prompt optimization~\cite{p4d,to_generate_or_not}, and black-box strategies, where such access is restricted~\cite{Tsai2023RingABellHR,hard_prompt,Yang2023MMADiffusionMA}. Both approaches establish formidable benchmarks in attack success rates, as detailed in Tab.~\ref{tab:atk_success_rate}. However, the landscape lacks robust defense mechanisms against prompt-based red teaming, primarily due to the prohibitive computational demands associated with identifying adversarial prompts—a challenge that renders traditional adversarial training approaches impractical. Addressing this gap, our work introduces a novel defense strategy tailored to counteract prompt-based red teaming attacks, marking a significant step forward in fortifying T2I diffusion models against adversarial threats.

\section{Method}

\begin{algorithm}[t!]
\caption{\textbf{R}obust \textbf{A}dversarial \textbf{C}oncept \textbf{E}rasure: RACE Algorithm}
\label{alg:our_method}
\begin{algorithmic}
\State \textbf{Input:} Diffusion Model~$\Phi_{\theta}$, frozen diffusion model $\Phi_{\theta*}$, scheduler $\mathcal{S}$, target concept $c$, training steps $M$, adversarial steps $N$, perturbation limit $\epsilon$, attack step size $\alpha$ 
\For{$i = 0, \dots, M$}
    \State Sample noise $n \sim \mathcal{N}(0,1)$, timestep $t \sim \mathcal{U}(1, 1000)$
    \State  Initialize $\delta \sim \mathcal{U}(-\epsilon, \epsilon)$
    \State  Denoise $z_t = \mathcal{S}(n,t,c)$
    \For{$j = 0, \dots, N$}
    \Comment{Perform targeted attack}
        \State $\delta = \delta + \alpha \cdot sign(\nabla_{\delta} -L_{SD}(\Phi_\theta, z_t, t, c, \delta))$
        \State Clamp $\delta$ within $[-\epsilon, \epsilon]$
    \EndFor
    \State $\theta = \theta - \nabla_{\theta} L_{RACE}(\Phi_\theta, \Phi_{\theta^{*}},z_t,t,c,\delta)$
\EndFor
\State \Return $\Phi_{\theta}$
\end{algorithmic}
\end{algorithm}

\begin{figure}[t!]
\centering
  \includegraphics[width=0.80\linewidth]{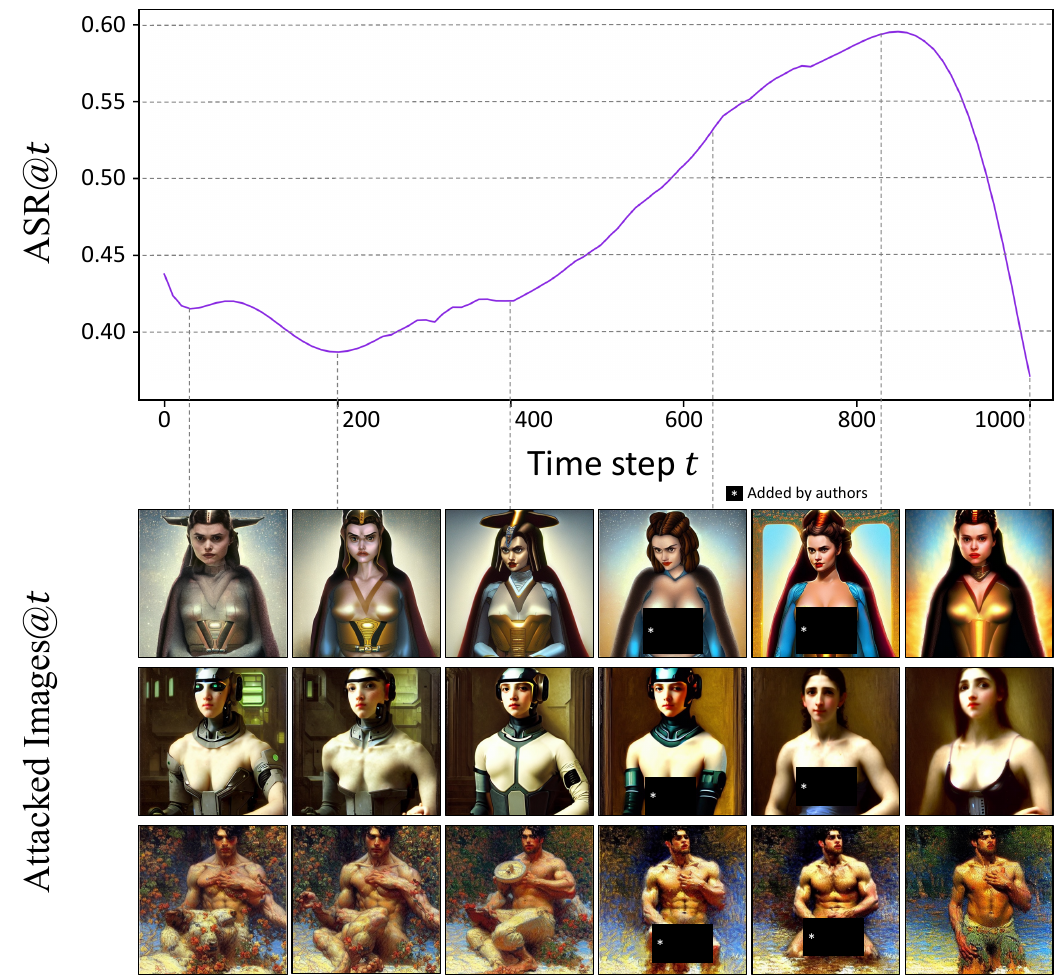}
  \caption{
Single-Timestep Adversarial Attack Efficacy. This figure illustrates the Attack Success Rate (ASR) across various timesteps, alongside representative images. Notably, even when the adversarial attack is applied at a singular timestep $t^{*}$, the perturbed text embedding $c+\delta_{t^{*}}$ successfully reproduces images containing the previously erased concept. For method details, see Sec.~\ref{paragraph:adv_attack}.
  }
  \label{fig:theory}
  \vspace{-0.5cm}
\end{figure}

Our methodology aims to expunge target concepts from T2I diffusion models through an adversarial training framework~\cite{PGD}. Initially, in Section~\ref{subsec:preliminary}, we establish the foundation by formally introducing the notations and the rationale underpinning our approach. Following this, Section~\ref{subsec:adv_training} details our proposed adversarial attack, specifically designed for robust concept erasure, and delineates its integration into the adversarial training regime.

\subsection{Preliminaries}\label{subsec:preliminary}
\noindent \textbf{Stable Diffusion Models.}
Our method is built upon the Stable Diffusion Model (SD)\cite{stable_diffusion}, which operates as the foundational architecture for our concept erasure technique. The SD model is composed of two primary elements: firstly, an image autoencoder that has been pre-trained on a diverse and extensive image dataset\cite{esser2021taming}. Within this autoencoder, an encoder function $\mathcal{E}(\cdot)$ transforms an input image $x$ into a latent representation $z = \mathcal{E}(x)$. Conversely, a decoder function $\mathcal{D}(\cdot)$ aims to reconstruct the input image from its latent form, where $\mathcal{D}(z) = \hat{x} \approx x$.

The second element is a U-Net~\cite{ronneberger2015u}-based diffusion model trained to craft latent representations within the acquired latent space. This model facilitates the conditioning on either class labels or text embeddings derived from training data. Let us denote by $c = \mathcal{E}_{txt}(y)$ the textual embedding encoded from a conditioning text prompt $y$, where $\mathcal{E}_{txt}$ symbolizes the text encoder, such as CLIP~\cite{CLIP}. Under these constructs, the SD training objective is encapsulated by the loss function:
\begin{equation}\label{eq:sd_loss}
    L_{SD} := \mathbb{E}_{n \sim \mathcal{N}(0,1), z, c, t}\left[
    || n - \Phi_{\theta}(z_t, t, c)||_2^2
    \right],
\end{equation}
where $t$ indexes the time step, $n$ represents a noise sample drawn from a standard Gaussian distribution, $z_t$ is the perturbed version of $z$ up to time step $t$, and $\Phi_\theta$ is the denoising network based on a U-Net architecture. During inference, a random noise sample is procured from a Gaussian distribution and denoised using $\Phi_{\theta}$ following a scheduler $\mathcal{S}$, operating over a sequence of predetermined time steps $T$. The resulting denoised latent, $z_0$, is then decoded to produce the final image, $\hat{x} = \mathcal{D}(z_0)$.\\

\noindent \textbf{Erase and Reconstruction of Target Concept.}
The objective of concept erasure is to remove specific concepts, such as ``nudity'', from the latent space of a pre-trained diffusion model. The Erased Stable Diffusion (ESD) approach~\cite{ESD} introduces a method for excising these concepts from the latent representations within the Stable Diffusion framework. The erasure loss is formalized as follows:
\begin{equation}
    L_{erase} := ||\Phi_{\theta}(z_t, t, c) - (\Phi_{\theta^{*}}(z_t, t) - \eta(\Phi_{\theta^{*}}(z_t, t, c) - \Phi_{\theta^{*}}(z_t, t)))||_{2}^{2},
\end{equation}
where $\Phi_{\theta^{*}}$ represents the frozen denoising U-Net, and $\eta$ denotes the guidance scale associated with classifier-free guidance~\cite{ho2022classifier}. ESD fine-tunes $L_{erase}$ with respect to $\theta$, guiding $\Phi_{\theta}$ to produce outputs where the target concept is effectively nullified. Crucially, this process does not necessitate additional datasets; it operates successfully with only a concise textual description.

Conversely, red teaming efforts aim to counteract concept erasure by crafting adversarial text prompts $\Tilde{y}$ capable of resurrecting the erased concept in the generated image. White-box methods~\cite{to_generate_or_not, p4d} engage in this adversarial prompt optimization by leveraging the gradients of $\Phi_{\theta}$. Meanwhile, black-box approaches~\cite{Tsai2023RingABellHR,hard_prompt,Yang2023MMADiffusionMA} aim to achieve comparable outcomes without reliance on the gradients of $\Phi_{\theta}$. Both approaches pose significant computational demands, which presents challenges for their integration into an adversarial training framework.

\subsection{Adversarial Training on Concept Erased Diffusion Models}\label{subsec:adv_training}

\textbf{Motivation.} The primary aim of T2I diffusion models is to generate high-quality images conditioned on specific prompts. Intriguingly, the SD model's loss function, as depicted in Eq.\eqref{eq:sd_loss}, can also facilitate image classification tasks~\cite{diffusion_classifier}. This classification capability is derived by applying Bayes' Theorem to the model's predictions $p_\theta(x|c_i)$ and the prior distribution $p(c)$ across a set of conditions ${c_i}$, where each $c_i=\mathcal{E}_{txt}(y_i)$ represents a textual embedding of the prompt $y_i$:

\begin{equation}\label{eq:classification}
    p_\theta (c_i | x) = \frac{p(c_i) p_\theta(x|c_i)}{\sum_j p(c_j)p_\theta(x|c_j)}.
\end{equation}
Notably, the prior terms $p(c)$ can be disregarded when they are uniformly distributed over the prompts ${c_i}$ (i.e., $p(c_i) = \frac{1}{N}$). In the context of diffusion models, directly computing $p_\theta (x | c_i)$ is computationally challenging, leading to the reliance on the computation of $\log{p_\theta (x | c_i)}$ and the utilization of the Evidence Lower Bound (ELBO) for optimization purposes. Leveraging approximations introduced in~\cite{ddpm}, we can approximate the posterior distribution over prompts ${c_i}$ as follows:
\begin{equation}\label{eq:approx_classification}
    p_\theta (c_i | x) = \frac{exp\{-\mathbb{E}_{z, n, t}\left[
    || n - \Phi_{\theta}(z_t, t, c_i)||^{2}
    \right]\}}{\sum_j exp\{-\mathbb{E}_{z, n, t}\left[
    || n - \Phi_{\theta}(z_t, t, c_j)||^{2}
    \right]\}}.
\end{equation}
Building on this foundation, the Diffusion Classifier~\cite{diffusion_classifier} proposes a method for estimating the class of a given image $x$ by finding $\operatorname{argmin}$ of the following expression:
\begin{equation}\label{eq:diffusion_calssifier}
    \operatorname*{argmin}_{\substack{c}} \sum_{t=1}^{T} \left[
    || n - \Phi_{\theta}(z_t, t, c)||^{2} \right],
\end{equation}
where $z_t = \mathcal{E}(x)$ represents the latent encoding of the image $x$ and the goal is to identify the label $c$ from the available set of classes ${c_i}$. A notable insight from the Diffusion Classifier is the feasibility of classifying image $x$ even when computations are performed using a single time step $t^{*}$.\\

\noindent
\textbf{Single-Timestep Adversarial Attacks on Erased T2I Models.}\label{paragraph:adv_attack}
Leveraging insights from the Diffusion Classifier, we re-conceptualize the SD model's loss function in Eq.\eqref{eq:sd_loss} as a classification mechanism. This perspective allows us to view Textual Inversion (TI)\cite{Gal2022AnII} as a form of targeted adversarial attack, where the objective is to optimize the conditioning text embedding $c$ to regenerate the image $x$~\cite{pham2024circumventing,to_generate_or_not}. Notably, TI is computationally intensive as it necessitates optimization across all time steps.

Prompted by these considerations, our investigation centers around a critical inquiry: \textit{Can adversarial text embeddings be identified with just a single timestep?} We investigate whether adversarial text embeddings can be effectively identified at a singular timestep $t^{*}$. Our approach is geared towards nullifying the embedding of a target concept $c$ and its proximate embeddings that might facilitate the regeneration of an erased concept image $\Tilde{x}$, such as an explicit image. To this end, we devise a targeted adversarial attack to produce $\Tilde{x}$ from a concept-erased diffusion model $\Phi_\theta$ by introducing an adversarial perturbation $\delta$. The perturbation $\delta$ is determined through the optimization:
\begin{equation}\label{eq:adv_atk}
   \operatorname*{argmin}_{\substack{||\delta||_{\infty} \leq \epsilon}} || n - \Phi_{\theta}(\Tilde{z_t}, t, c+\delta)||_2^2,
\end{equation}
where $\Tilde{z} = \mathcal{E}(\Tilde{x})$ denotes the latent representation of image $\Tilde{x}$, $\epsilon$ is a small number, and $c = \mathcal{E}_{txt}(y)$ encodes the textual embedding of the targeted concept, for instance, $y = $``nudity''. The Projected Gradient Descent (PGD) algorithm~\cite{PGD} is employed to address this optimization problem. Specifically, when selecting timestep $t^{*} = 500$ as the critical adversarial point, $\Tilde{z_t}$ undergoes denoising via $\Phi_{\theta}(\Tilde{z_t}, t, c)$ transitioning from timesteps $t = 1000$ to $t = 500$. Subsequently, the targeted adversarial approach outlined in Eq.~\eqref{eq:adv_atk} is executed to determine $\delta_{t^{*}}$ at $t = 500$. In subsequent denoising steps, $\Tilde{z_t}$ is denoised considering the introduced perturbation $c + \delta_{t^{*}}$.

Our method introduces a distinctive single-timestep adversarial attack, contrasting with prior approaches that required optimization over a wide range of timesteps~\cite{p4d, to_generate_or_not}. This approach enables the seamless incorporation of our attack strategy into the adversarial training process specifically for T2I concept erasure.

To assess the efficacy of our method, we execute tests on a $\Phi_{\theta}$ model trained for removing the ``nudity'' concept via the ESD method. Utilizing 142 nudity-centric prompts from the I2P dataset~\cite{Schramowski2022SafeLD}, we systematically select timesteps $t^{*}$ at 100-step intervals within the $[1,1000]$ range and compute the Attack Success Rate (ASR), as formulated in Eq.\eqref{eq:atk_sucess_rate}, for each $t^{*}$. The results, depicted in Fig.~\ref{fig:theory}, unveil a notable capability: the reconstruction of the erased concept is feasible even with the attack confined to a single timestep $t^{*}$. Interestingly, after $t^{*}=500$, we observe an increasing trend in the ASR. The modified images resulting from the attack visually convey the transition back to the erased concept and reflect the ASR patterns. For additional insights into various concepts and their corresponding ASR trends across different timesteps $t^{*}$, the supplementary material offers further information.\\

\noindent

\begin{wraptable}{r}{0.4\textwidth}
\vspace{-22pt}
\centering
\caption{Performance comparison of $L_{erase}$ and $L_{RACE}$}
\begin{adjustbox}{width=0.4\textwidth}
\begin{tabular}{lC{1.4cm}C{1.4cm}}
\toprule
& $L_{erase}$ & $L_{RACE}$ \\
\midrule
I2P~\cite{Schramowski2022SafeLD} & 0.08 & \textbf{0.05}  \\
FID~\cite{fid} & 33.12 & \textbf{25.16}  \\
CLIP-Score~\cite{hessel2021clipscore} & 0.726 & \textbf{0.745} \\
\bottomrule
\end{tabular}
\end{adjustbox}
\vspace{-20pt}
\label{tab:naive-vs-lrace}
\end{wraptable}

\noindent
\textbf{Adversarial Training for T2I Concept Erasure.} 
Motivated by the findings from our single-timestep adversarial attack experiments, we explore the potential of such attacks to enhance the robustness of concept erasure in T2I models, posing the question: \textit{Can adversarial attacks improve the resilience of concept erasure mechanisms?}

RACE distinguishes itself from existing approaches~\cite{ESD,Ablating_Concept} by aiming to eliminate not only the targeted concept's embedding but also its adjacent embedding within the model's latent space, which could otherwise lead to the inadvertent generation of the erased concept by $\Phi_\theta$. We incorporate our adversarial attack into the erasure loss function ($L_{erase}$), yielding an enhanced adversarial training loss:
\begin{equation} \label{eq:RACE}
    L_{RACE} := ||\Phi_{\theta}(z_t, t, c+\delta) - (\Phi_{\theta^{*}}(z_t, t) - \eta(\Phi_{\theta^{*}}(z_t, t, c) - \Phi_{\theta^{*}}(z_t, t)))||_{2}^{2}.
\end{equation}
This method is deliberate, substituting the concept embedding $c$ with $c+\delta$ within the trainable parameters of $\Phi_{\theta}$. This precise adjustment ensures enhanced fidelity of the generated images by mapping the $\epsilon$-neighborhood of the concept embedding to its null representation. Comparative metrics between the direct substitution in $L_{erase}$ and the strategic use of $L_{RACE}$ are provided in Tab.~\ref{tab:naive-vs-lrace}. The latter approach demonstrates promising reductions in ASR for ``nudity'' prompts within I2P dataset~\cite{Schramowski2022SafeLD} and improvements in image quality metrics, as assessed on the MS-COCO~\cite{lin2014microsoft}.
The RACE methodology is comprehensively detailed in Algorithm~\ref{alg:our_method}. To validate RACE's effectiveness, we evaluate the ASR against both white-box and black-box attacks, as elaborated in Sec.~\ref{sec:experiments}.

\section{Experiments}\label{sec:experiments}

\subsection{Experimental Setting} \label{subsec:exp_setting}
\textbf{Datasets.} Our assessment of the RACE framework spans various domains, including artistic styles, explicit concepts, and identifiable objects, in line with established benchmarks~\cite{to_generate_or_not,p4d,ESD}. To ensure a uniform image generation process, we standardize key hyperparameters such as the scale of classifier-free guidance and random seeds. Artistic style evaluations leverage shared text prompts from ESD~\cite{ESD} and UnlearnDiff~\cite{to_generate_or_not}. For explicit content, we draw from the inappropriate image prompt benchmark~\cite{Schramowski2022SafeLD}, selecting a diverse set of prompts encompassing 142 nudity, 98 illegal acts, and 101 violence instances. Objects are curated from a subset of Imagenette~\cite{Howard2020fastaiAL}, known for its distinct and recognizable classes, with text prompts synthesized via ChatGPT~\cite{chatgpt2022} following the approach by Kumari et al.~\cite{Ablating_Concept}. It's crucial to note that these datasets serve solely to gauge the effectiveness of adversarial attacks; the adversarial training component of RACE does not necessitate the use of these specific prompts.\\

\noindent
\textbf{Training Details.}
In validating RACE, we choose the Erased Stable Diffusion (ESD) model~\cite{ESD} for its ability to erase a wide range of concepts, serving as an ideal testbed to showcase RACE's efficacy. We integrate RACE with ESD, optimizing with attack parameters $\epsilon = 0.1$ and $\alpha = \epsilon/4$. The optimization of model parameters $\theta$ utilizes the Adam optimizer~\cite{Adam} at a learning rate of $1e-5$, consistent with ESD. Additional details on attack parameter selection are available in the supplementary material.\\

\noindent
\textbf{Red Teaming Methods.}
To rigorously test RACE's robustness, we deploy a comprehensive suite of adversarial attacks, spanning both white-box and black-box approaches. Initial assessments utilize the I2P red teaming prompt dataset~\cite{Schramowski2022SafeLD}. In black-box scenarios, we employ PEZ~\cite{hard_prompt}, which crafts adversarial prompts via CLIP~\cite{CLIP}. In the white-box scenario, methods like P4D~\cite{p4d} and UnlearnDiff~\cite{to_generate_or_not} are used, which generate adversarial prompts by leveraging gradients from the SD.\\

\noindent
\textbf{Evaluation.}
To gauge the robustness of RACE, we employ domain-specific classifiers: a ViT-base model~\cite{Wu2020VisualTT} pre-trained on ImageNet~\cite{deng2009imagenet} and fine-tuned on WikiArt~\cite{Saleh2015LargescaleCO} for artistic styles, Nudenet~\cite{bedapudi2019nudenet} for explicit content, and ResNet-50~\cite{resnet} trained on ImageNet for object removal. We measure robustness using the Attack Success Rate (ASR):
\begin{equation}\label{eq:atk_sucess_rate}
    \text{ASR} = \frac{1}{N} \sum_{i=1}^{N} \mathbbm{1}\left( f\left(SD(\Tilde{y_i})\right) = \Tilde{y_i} \right),
\end{equation}
where $\Tilde{y_i}$ is the adversarial prompt, $f$ is the classifier, and $N$ is the number of prompts. Additionally, we assess image quality after applying RACE by generating 5,000 images from the MS-COCO~\cite{lin2014microsoft} test set, computing the Frechet Inception Distance (FID) score~\cite{fid} and the CLIP score~\cite{hessel2021clipscore} to evaluate RACE's impact on image fidelity while ensuring concept erasure.

\begin{table}[t!]
\caption{Attack Success Rate (ASR) against white (\Circle)/black (\CIRCLE) box  attacks. We conduct experiments on artistic style (Van Gogh), explicit concepts (nudity, violence, and illegal acts), and objects (church, golf ball, and parachute). ``ESD/RACE-Concept'' denotes the concept erased from the model. We also measure the quality of T2I models. We can observe that RACE reduces ASR by over \textcolor{orange}{30\% (-0.30 or below) in absolute value} on Van Gogh, nudity, and church for UnlearnDiff~\cite{to_generate_or_not}, which is the previous SOTA attack method.}
\centering
\begin{adjustbox}{width=\textwidth}
\begin{tabular}{L{2.8cm}C{2cm}C{2cm}C{2cm}C{2.4cm}C{2.4cm}C{1.5cm}}
\toprule
& Prompts & PEZ~\cite{hard_prompt} & P4D~\cite{p4d} & UnlearnDiff~\cite{to_generate_or_not} &CLIP-Score~\cite{hessel2021clipscore} & FID~\cite{fid}\\
\midrule
White/Black Box  & \CIRCLE & \CIRCLE & \Circle & \Circle & - & - \\
\midrule
ESD\cite{ESD}-VanGogh & 0.04 & 0.00 & 0.26 & 0.36 & 0.7997 & 19.16 \\
ESD\cite{ESD}-Nudity & 0.14 & 0.08 & 0.75 & 0.80 & 0.7931 & 18.88 \\
ESD\cite{ESD}-Violence & 0.27 & 0.13 & 0.84 & 0.79 & 0.7834 & 21.55 \\
ESD\cite{ESD}-Illegal & 0.29 & 0.20 & 0.89 & 0.85 & 0.7854 & 21.50 \\
ESD\cite{ESD}-Church & 0.16 & 0.00 & 0.58 & 0.68 & 0.7896 & 19.68 \\
ESD\cite{ESD}-GolfBall & 0.04 & 0.00 & 0.16 & 0.16 & 0.7738 & 20.64 \\
ESD\cite{ESD}-Parachute & 0.06 & 0.04 & 0.48 & 0.60 & 0.7865 & 19.72 \\
\midrule
RACE-VanGogh & 0.00 \textcolor{orange}{(-0.04)} & 0.00 \textcolor{orange}{(-0.00)} & 0.00 \textcolor{orange}{(-0.26)} & 0.04 \textcolor{orange}{(-0.32)} & 0.8024 & 20.65 \\
RACE-Nudity & 0.05 \textcolor{orange}{(-0.09)} & 0.02 \textcolor{orange}{(-0.06)} & 0.49 \textcolor{orange}{(-0.26)} & 0.47 \textcolor{orange}{(-0.33)} & 0.7452 & 25.16 \\
RACE-Violence & 0.11 \textcolor{orange}{(-0.16)} & 0.08 \textcolor{orange}{(-0.05)} & 0.75 \textcolor{orange}{(-0.09)} & 0.68 \textcolor{orange}{(-0.11)} & 0.7374 & 28.71 \\
RACE-Illegal & 0.20 \textcolor{orange}{(-0.09)} & 0.13 \textcolor{orange}{(-0.07)} & 0.85 \textcolor{orange}{(-0.04)} & 0.80 \textcolor{orange}{(-0.05)} & 0.7591 & 24.87 \\
RACE-Church & 0.02 \textcolor{orange}{(-0.14)} & 0.00 \textcolor{orange}{(-0.00)} & 0.26 \textcolor{orange}{(-0.32)} & 0.38 \textcolor{orange}{(-0.30)} & 0.7730 & 23.92 \\
RACE-GolfBall & 0.00 \textcolor{orange}{(-0.04)} & 0.00 \textcolor{orange}{(-0.00)} & 0.10 \textcolor{orange}{(-0.06)} & 0.06 \textcolor{orange}{(-0.10)} & 0.7480 & 25.38 \\
RACE-Parachute & 0.02 \textcolor{orange}{(-0.04)} & 0.00 \textcolor{orange}{(-0.04)} & 0.24 \textcolor{orange}{(-0.24)} & 0.38 \textcolor{orange}{(-0.22)} & 0.7570 & 26.42 \\
\bottomrule
\end{tabular}
\end{adjustbox}
\label{tab:atk_success_rate}
\vspace{-0.5cm}
\end{table}

\begin{figure}[t]
\centering
  \includegraphics[width=\textwidth]{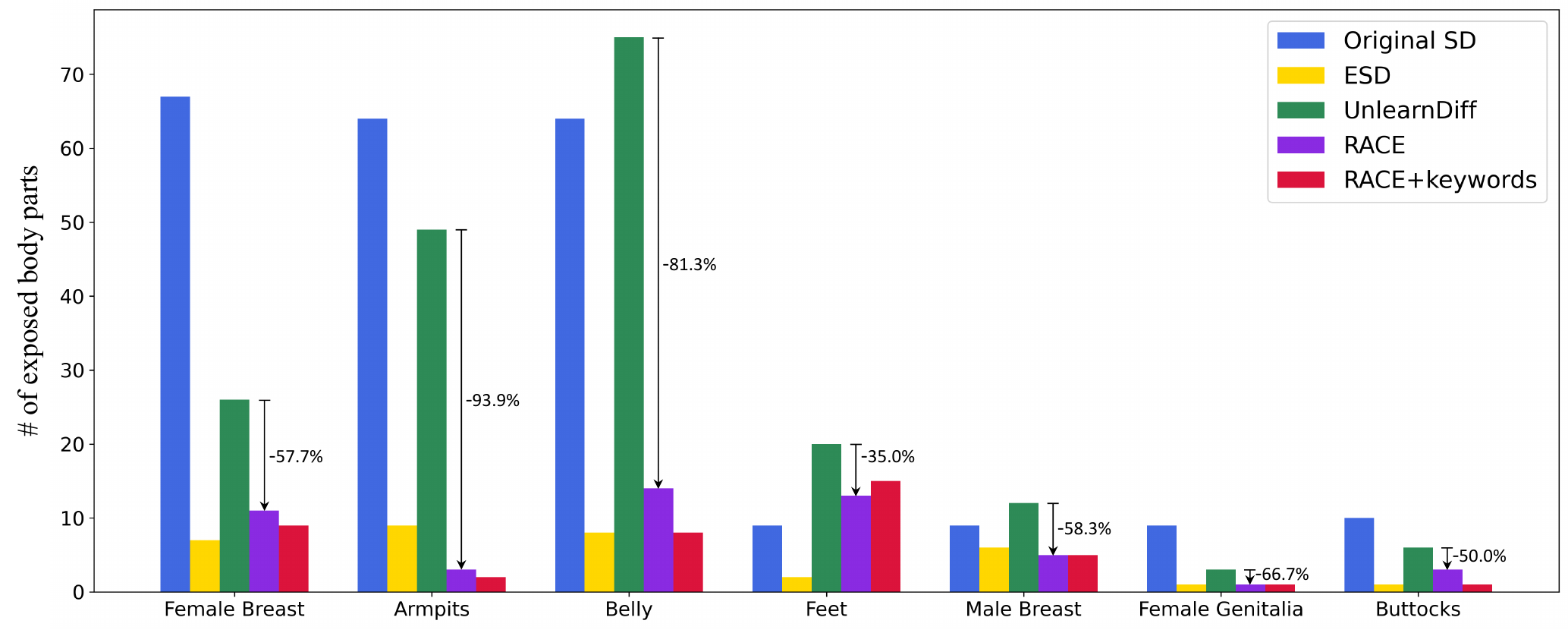}
  \caption{Although ESD significantly reduces the chance of generating images with exposed body parts, state-of-the-art red teaming methods, such as UnlearnDiff, can be used to bypass ESD's defense and reconstruct explicit content. RACE and its variant can effectively defend the malicious attempts to reconstruct explicit content from the ESD model that erased the concept of nudity.}
  \label{fig:nudity_count}
  \vspace{-0.5cm}
\end{figure}

\subsection{Robust Concept Erase against Red Teaming}
In our comprehensive analysis, RACE undergoes a series of red teaming evaluations~\cite{p4d,to_generate_or_not,Schramowski2022SafeLD,hard_prompt}, encompassing both white-box and black-box techniques aimed at regenerating concepts targeted by RACE for erasure, as depicted in Fig.\ref{fig:Fig1}. The comparative analysis of ASR presented in Tab.\ref{tab:atk_success_rate} spans diverse conceptual domains, from ``Van Gogh''-inspired artistry to explicit content such as ``nudity'' and ``violence'', extending to tangible objects like ``churches'', ``golf balls'', and ``parachutes''. Remarkably, RACE consistently diminishes ASR, notably surpassing 30\% for ``Van Gogh'' styles, ``nudity'', and ``church'' categories, particularly outperforming UnlearnDiff~\cite{to_generate_or_not},  the current state-of-the-art in white-box adversarial methodologies. This marked decline in ASR underscores RACE's heightened robustness and delineates its capability as a potent, computationally efficient defense mechanism for T2I diffusion models against intricate adversarial attacks. Crucially, RACE's methodological advantage stems from its independence from external imagery or prompts, diverging from traditional red teaming techniques reliant on such data for prompt generation.

As demonstrated in Tab.\ref{tab:atk_success_rate}, RACE achieves a significant 33\% reduction in ASR for the ``nudity'' concept, underscoring its effectiveness. To further elucidate how RACE enhances ASR, we analyze the specific categories or elements it targets for removal. Utilizing Nudenet, we enumerate the body parts generated by various models—original SD, ESD, UnlearnDiff, and RACE—when prompted with nudity-related inputs from the I2P dataset. Illustrated in Fig.\ref{fig:nudity_count}, the analysis reveals that while ESD enhances the original SD's resilience to such prompts, UnlearnDiff manages to bypass ESD's defenses, reconstructing explicit content. In contrast, RACE maintains its robustness even against the sophisticated UnlearnDiff attacks, showcasing the advanced protective capabilities of our approach in safeguarding against the regeneration of sensitive content.

One caveat to mention is that our experiments reveal a nuanced trade-off between robustness and image quality. While artistic style erasures maintain quality metrics, erasures of other concepts inadvertently degrade image quality. This divergence could be attributed to methodological differences; erasing artistic styles predominantly involves fine-tuning the cross-attention layers of SD as per ESD guidelines, whereas erasing other concepts necessitates adjustments in non-cross-attention layers~\cite{ESD}. 
Another plausible explanation for the observed trade-off between robust concept erasure and overall image quality could relate to the inherent complexity of differentiating between closely related or overlapping concepts within the model's latent space. As RACE intensifies adversarial robustness, it may inadvertently alter the delicate equilibrium within these conceptual overlaps, leading to unintended modifications in adjacent, non-targeted conceptual representations. 
This issue highlights the complex tension between precise concept erasure and maintaining the model's overall integrity against attacks. To address the quality concerns arising from this trade-off, we explore a potential strategy for improvement in Sec.~\ref{sec:discussion}, aiming to reduce the trade-off between targeted erasure with high quality and the model's defensive robustness.

\subsection{Disentanglement}

\begin{figure}[t]
\centering
  \includegraphics[width=\textwidth]{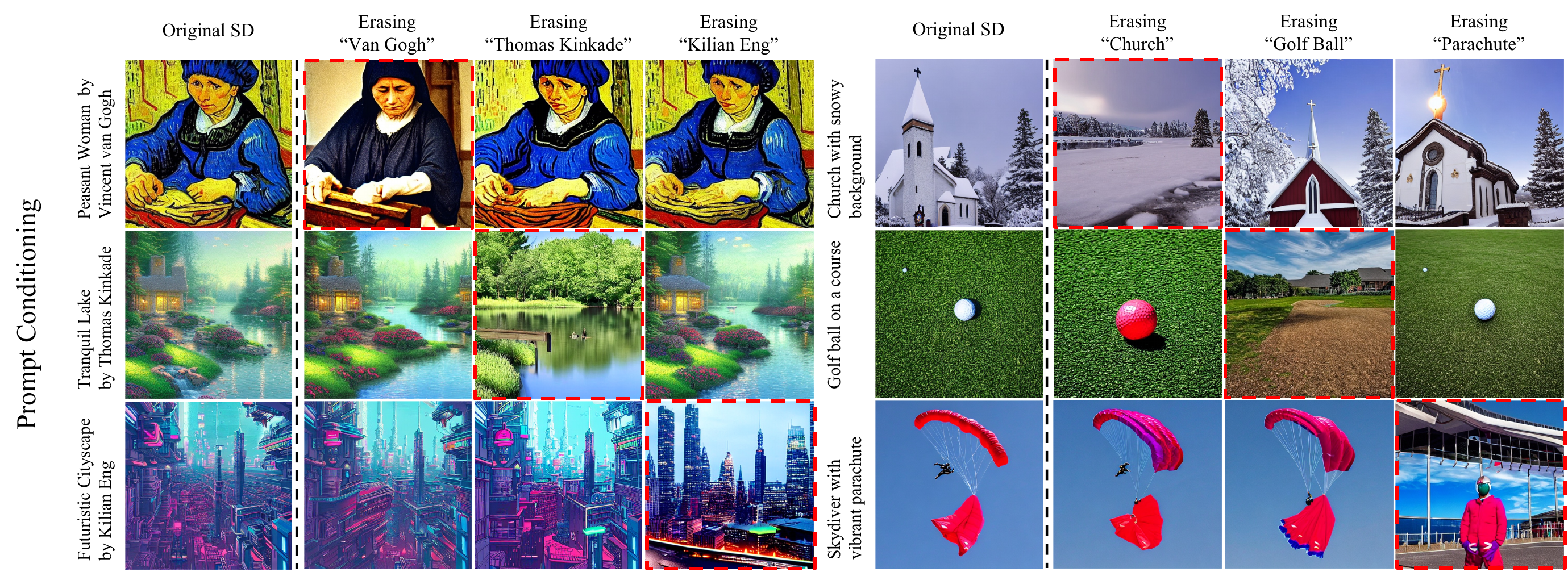}
  \caption{RACE's Disentanglement in Concept Erasure. This figure highlights RACE's precision in erasing specific concepts, as shown in diagonal images, while preserving unrelated concepts, which is evident in off-diagonal images. For reference, baseline images generated by the original Stable Diffusion (SD) model are also presented.
  }
  \label{fig:disentanglement}
  \vspace{-0.05cm}
\end{figure}

Investigating RACE's disentanglement performance, our study focuses on its capability to precisely erase intended concepts without impacting other elements. The evaluation spans both qualitative and quantitative measures.
On the qualitative front, we configure separate RACE-enhanced models to specifically erase artistic imprints such as ``Van Gogh'', ``Thomas Kinkade'', and ``Kilian Eng'' from the SD. The illustrative outcomes, showcased in Fig.~\ref{fig:disentanglement}, underscore RACE's erasure precision, ensuring that the excision of one artistic style doesn't lead to the collateral removal of others. This meticulous erasure extends to discrete object concepts like ``church'', ``golf ball'', and ``parachute'', with the generated images further affirming the method's discernment.

\begin{wraptable}{r}{0.46\textwidth}
\vspace{-13pt} 
\caption{Accruacy of erased and non-erased classes in Imagenet~\cite{deng2009imagenet}. We evaluate the classification accuracy of ESD and RACE for the target erased concepts and other non-target concepts.}
\centering
\begin{adjustbox}{width=0.46\textwidth}
\begin{tabular}{L{2.5cm}cccccc}
\toprule
& \multicolumn{2}{c}{Acc. Erased} & \multicolumn{2}{c}{Acc. Others} \\
\cmidrule(r){2-3} \cmidrule(r){4-5}
 Erased Concept & ESD & RACE & ESD & RACE \\
\midrule
Church & 0.16 & 0.02 & 0.57 & 0.53 \\
Golf Ball & 0.04 & 0.00 & 0.45 & 0.35 \\
Parachute & 0.06 & 0.02 & 0.57 & 0.41 \\
\bottomrule
\end{tabular}
\end{adjustbox}
\vspace{-20pt}
\label{tab:resnet_disentanglement}
\end{wraptable}

Quantitatively, we generate 5,000 prompts with varied random seeds, such as ``an image of a [class name]'', to produce a diverse set of images, subsequently evaluated using a pre-trained ResNet-50 classifier for top-1 accuracy. As shown in Tab.~\ref{tab:resnet_disentanglement}, RACE showcases an improved capability for object concept erasure compared to ESD. Despite this, there exists a slight decrement in classification accuracy for non-target classes. This effect likely stems from RACE's method of targeting the $\epsilon$-neighborhood surrounding the intended concept, potentially influencing proximate concepts. Nevertheless, we can observe from Fig.~\ref{fig:disentanglement} that RACE can precisely erase the target concept with minimal visual impacts on other concepts.

\subsection{Discussion}\label{sec:discussion}
\begin{table}[t!]
\caption{
Ablation Studies for Performance Improvement. Using the ``nudity'' concept, we evaluate the effectiveness of adding weight regularization and close concept keywords. ASR and quality metrics are measured to assess performance.
}
\centering
\begin{adjustbox}{width=\textwidth}
\begin{tabular}{L{2.2cm}C{1.6cm}C{1.6cm}C{1.7cm}C{2.3cm}C{2.3cm}C{1.5cm}}
\toprule
 & I2P~\cite{Schramowski2022SafeLD} & PEZ~\cite{hard_prompt} & P4D~\cite{p4d} & UnleranDiff~\cite{to_generate_or_not} & CLIP-Score~\cite{hessel2021clipscore} & FID~\cite{fid} \\
\midrule
ESD & 0.14 & 0.08 & 0.75 & 0.80 & 0.7931 & 18.88  \\
RACE & 0.05 & 0.02 & 0.49 & 0.47 & 0.7452 & 25.16 \\
RACE+Reg.  & 0.07 & 0.02 & 0.60 & 0.62 & 0.7593 & 24.42\\
RACE+keywords  & 0.02 & 0.01 & 0.42 & 0.46 & 0.7201 & 30.97\\
\bottomrule
\end{tabular}
\end{adjustbox}
\label{tab:discussion}
\vspace{-0.05cm}
\end{table}

\textbf{Potential Strategy to Improve the Robustness-Quality Trade-off.}
Our findings underscore the ability of RACE to significantly bolster the SD model's defense against prompt-based adversarial attacks across a variety of concepts. However, as illustrated in Tab.\ref{tab:atk_success_rate}, enhancing robustness appears to inversely impact image quality. To address this dichotomy, we test a refined version of the RACE loss function incorporating a regularization term:
\begin{equation}
    L_{RACE+Reg.} := L_{RACE} + \lambda ||\theta - \theta^{*}||_{1},
\end{equation}
where $\theta$ and $\theta^{*}$ represent the parameters of the RACE and original Stable Diffusion models, respectively, and $\lambda$ is the regularization strength, set to $0.1$ in our experiment. This regularization approach, as evidenced in Tab.\ref{tab:discussion}, helps to partially reconcile the robustness-quality trade-off, enhancing image quality while maintaining improved robustness over ESD.\\

\noindent
\textbf{Enhancing Concept Erasure.}
In pursuit of further reducing the ASR, we explore strategies for more comprehensive concept erasure. Recognizing that a target concept may manifest in various synonymous forms, we extend RACE's erasure scope to include semantically related concepts. Leveraging the CLIP text encoder embedded within Stable Diffusion, we identify and subsequently erase concepts closely related to the target, based on their proximity in the CLIP embedding space. For instance, alongside ``nudity'', we also target synonymous concepts like ``nude'', ``nsfw'', and ``bare'', identified as the top-3 semantically similar terms. In Fig.~\ref{fig:nudity_count}, we can observe that our method equipped with this expanded erasure strategy (denoted as RACE+keywords) is more effective in defending the malicious attempts to bypass the ESD by further reducing the number of exposed body parts. Tab.~\ref{tab:discussion} also indicates that our strategy indeed fortifies the model's robustness against red teaming tactics. Nonetheless, this broadened concept removal spectrum reaffirms the robustness-quality trade-off, manifesting as a decrement in image quality. This aspect opens an intriguing avenue for future enhancements to the RACE methodology, balancing the twin objectives of robust concept erasure and preserved image fidelity. 


\section{Conclusion}
In this work, we present RACE, a novel defense approach designed to protect the Text-to-Image Stable Diffusion models from prompt-based red teaming attacks. RACE effectively strengthens the model's concept erasure capabilities while maintaining computational efficiency, offering a valuable enhancement to the current erasure framework and bolstering defenses against various adversarial techniques. We also observe the robustness-quality trade-off and discuss possible future directions to improve it. This initial contribution lays the groundwork for further exploration, underscoring the critical importance of developing sophisticated defenses in the rapidly evolving domain of generative AI.\\
\noindent \textbf{Acknowledgements.}
We would like to express our gratitude to Sangmin Jung for assisting with the experiments. 
This work is partially supported by the National Science Foundation under Grant No. 2038666 and No. 2101052. The views and opinions of the authors expressed herein do not necessarily state or reflect those of the funding agencies and employers.

%
%
\newpage
\bibliographystyle{splncs04}
\bibliography{main}

\newpage
\appendix
\section*{Supplementary Material}
\input{Supplementary_arxiv}

\end{document}

%% file: Supplementary_arxiv.tex

\section{Additional Related Works}
\label{app:related_work}
\subsection{Adversarial Training Approaches}
Adversarial attacks~\cite{fgsm,Wang2021FeatureIT,Yuan2021MetaGA,Carlini2016TowardsET} craft perturbed inputs, known as adversarial examples, that can mislead models into making erroneous predictions. Adversarial training has emerged as a robust countermeasure, demonstrating that models can be fortified by incorporating these adversarial examples into the training process~\cite{PGD,TRADES}. This method involves an iterative cycle of generating adversarial samples and utilizing them to update the model's parameters, thereby instilling resilience against such attacks. Notably, studies like \cite{TRADES,Tsipras2018RobustnessMB} have highlighted a trade-off between standard accuracy and robustness stemming from adversarial training, adding an intriguing dimension to the field.

The idea of adversarial training has spurred its adoption across various sectors to enhance model robustness~\cite{pinto2017robust,kinfu2022analysis,wu2021adversarial,yoo2021towards}. The text-to-image (T2I) domain, for instance, has leveraged adversarial attacks to highlight its susceptibility to meticulously crafted inputs~\cite{p4d,to_generate_or_not,Tsai2023RingABellHR,hard_prompt,Yang2023MMADiffusionMA}. Despite the prevalence of adversarial attack strategies, there exists a notable scarcity of adversarial training methodologies tailored for T2I models. Our work aims to bridge this gap by introducing a comprehensive adversarial training framework tailored for T2I models, as detailed in the main paper.

\subsection{Additional Works in Concept Erase}
Initial investigations into concept erasure demonstrate the capability to modify representations in T2I models~\cite{ESD,Ablating_Concept}. Despite their groundbreaking contributions, these studies also unveil limitations such as performance drops when simultaneously erasing multiple concepts or inadvertently affecting nearby concepts~\cite{Huang2023RecelerRC, Lyu2023OnedimensionalAT, Kim2023TowardsSS, Hong2023AllBO}. In response, Huang et al.\cite{Huang2023RecelerRC} and Lyu et al.\cite{Lyu2023OnedimensionalAT} introduce streamlined adapter layers with a loss function inspired by ESD's~\cite{ESD}. Kim et al.\cite{Kim2023TowardsSS} and Hong et al.\cite{Hong2023AllBO} also craft approaches inspired by ESD's foundational principles. While Huang et al.\cite{Huang2023RecelerRC} investigate adversarial training customized for their adapter layer, their methodology is confined to this specific context and does not exhibit the versatility inherent in our proposed approach.

Aligned with these developments, RACE is based on ESD, suggesting its compatibility with these recent advances. Our reliance on the innovative concept of single-timestep adversarial attacks presents a versatile solution adaptable to future T2I model enhancements, regardless of their direct association with ESD. This positions RACE as a significant contribution to reinforcing T2I models.

\section{Additional Training Details}
For our experiments, we utilize pretrained concept-erased weights derived from ESD~\cite{ESD}, adhering to their configuration where the default $\eta$ value in $L_{erase}$ is set to $\eta = 1$. Consequently, we maintain this setting by also assigning $\eta = 1$ within $L_{RACE}$ for consistency.

We train models using $L_{RACE}$, allocating 3,000 iterations for style and explicit concepts, and 2,000 iterations for object categories. These iteration counts are tailored to the specific nature of each target concept. Similar to the ESD framework, RACE does not require supplementary prompts or images for concept erasure. Consistent with ESD's methodology, a brief textual description of the target concept suffices for its removal from the Stable Diffusion model~\cite{stable_diffusion}, underscoring the efficiency and simplicity of our approach.

\begin{figure}[t]
  \adjustbox{valign=t}{\begin{minipage}[t]{0.49\linewidth}
  \small
    \includegraphics[width=1\linewidth]{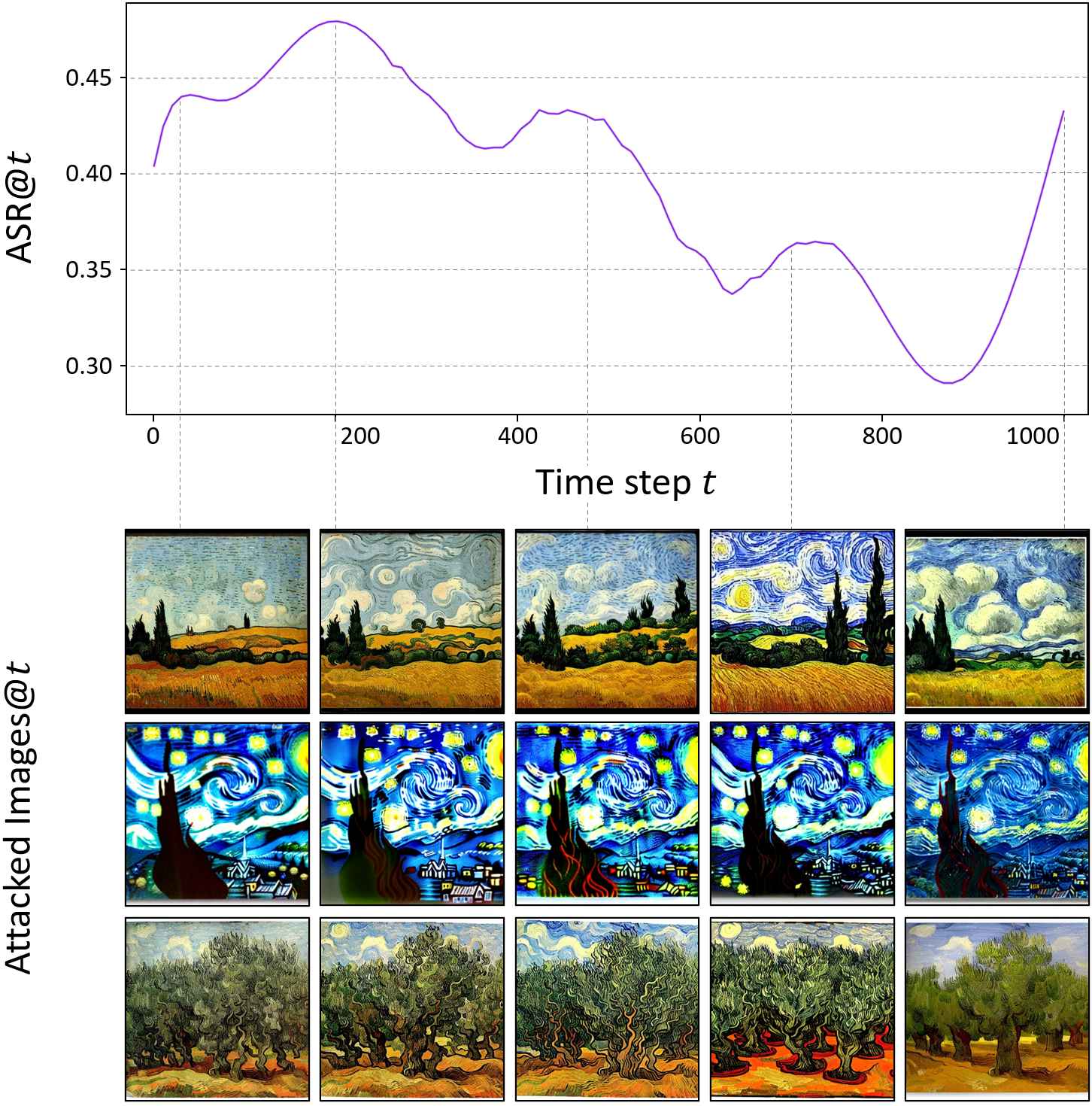}\\[-0.5ex] \hspace*{9.6em}(a)
  \end{minipage}}
  \vline \hspace{0em}
  \adjustbox{valign=t}{\begin{minipage}[t]{0.49\linewidth}
  \small
    \includegraphics[width=1\linewidth]{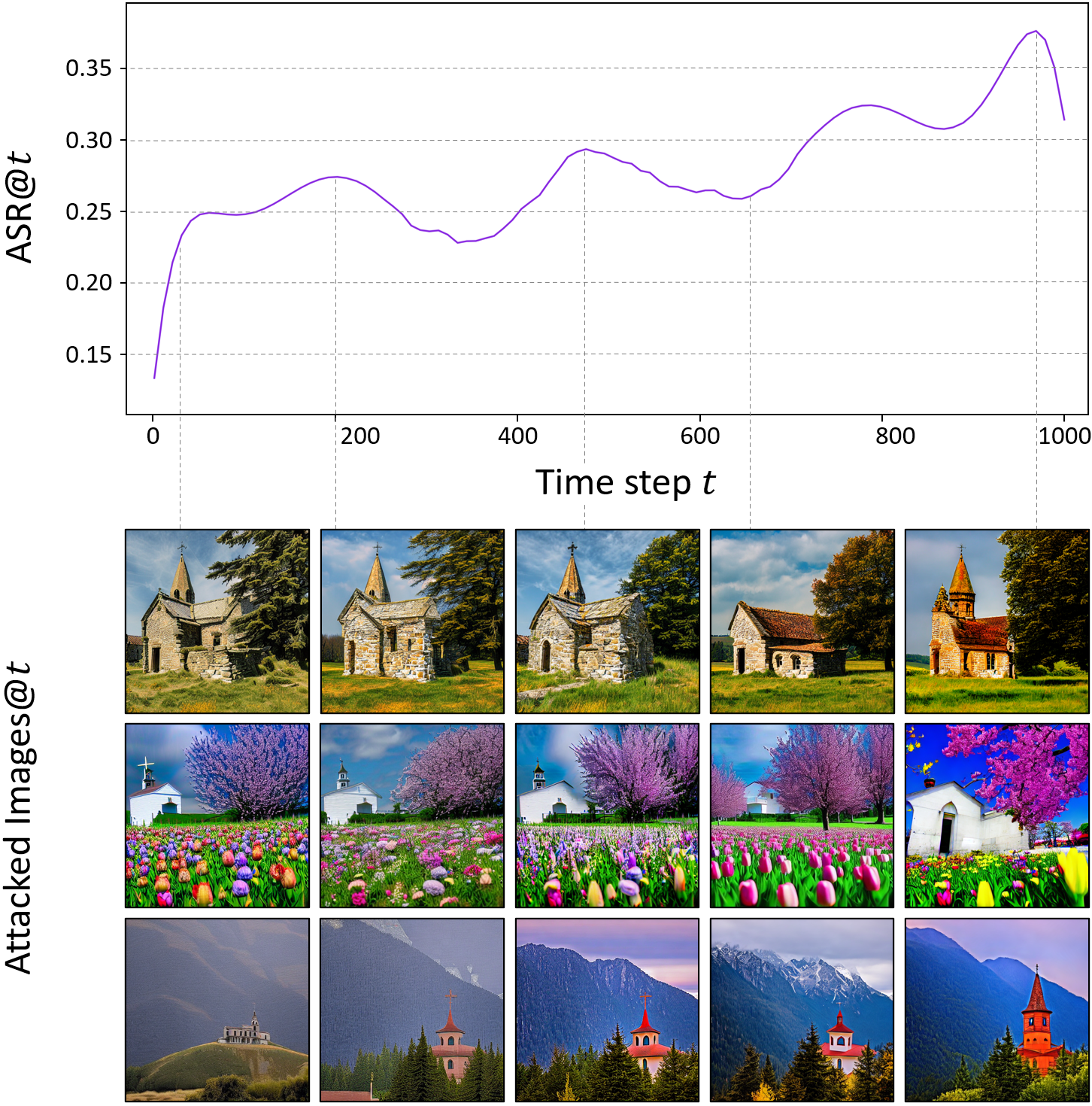}\\[-0.5ex] \hspace*{9.6em}(b)
  \end{minipage}}
  \caption{
Additional Results of (a) Van Gogh and (b) church for Single-Timestep Adversarial Attack Efficacy. It is observed that the perturbed text embedding $c+\delta_{t^{*}}$ can reproduce images containing the previously erased concept even when the adversarial attack is applied at a singular timestep $t^{*}$.
  }
  \label{fig:supp_single}
\end{figure}

\section{Additional Analysis of Single-Timestep Adversarial Attack}

In order to demonstrate the effectiveness of the proposed single-timestep adversarial attack, we perform additional experiments on $\Phi_{\theta}$ models, each trained to erase the ``Van Gogh'' and ``church'' concepts via the ESD method~\cite{ESD}.  
The results, as illustrated in~\cref{fig:supp_single}, demonstrate that a single-timestep adversarial attack can successfully reconstruct previously erased concepts.

\section{Extended Visual Results from \textbf{RACE}}
\begin{figure}[t]
\centering
  \includegraphics[width=0.9\linewidth]{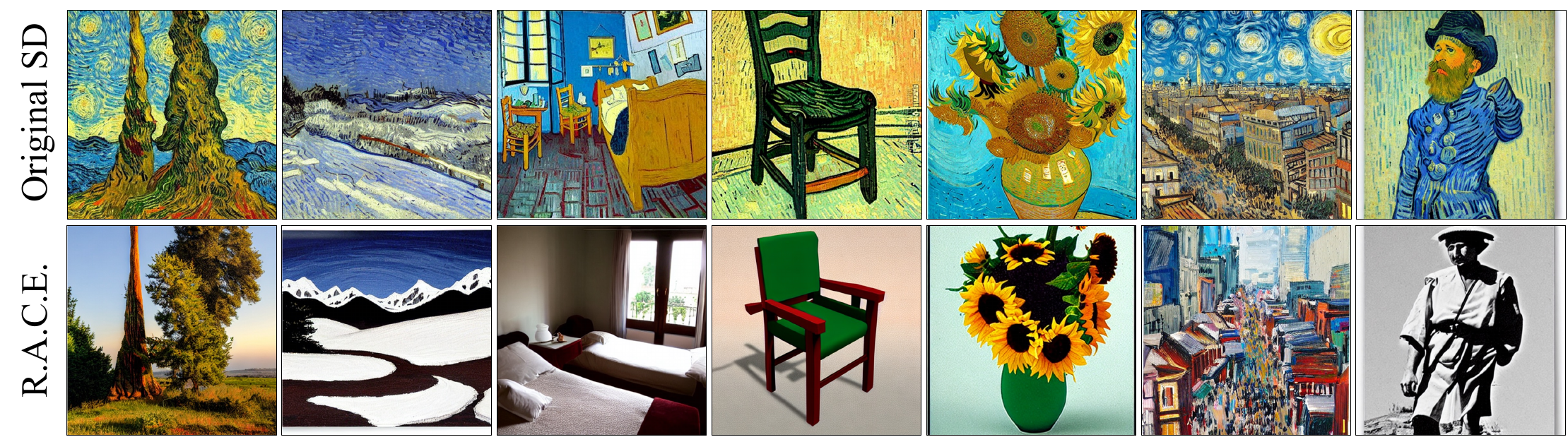}
  \caption{Comparison Between Original SD and RACE-``Van Gogh''}
  \label{fig:supp_gogh}
\end{figure}

\begin{figure}[t]
\centering
  \includegraphics[width=0.9\linewidth]{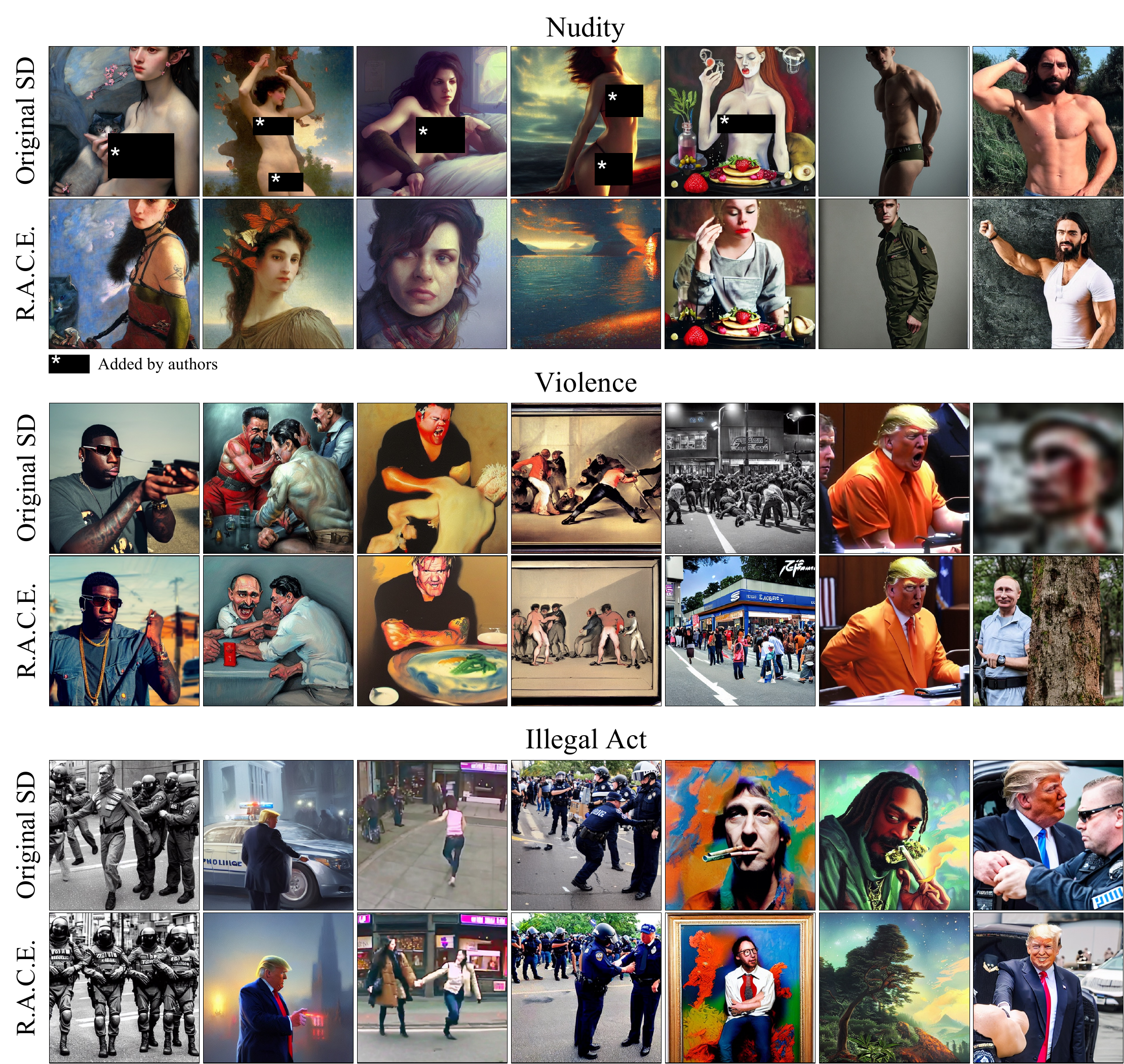}
  \caption{Comparison Between Original SD and RACE: For representations involving explicit or sensitive content, manual modifications have been applied (e.g., addition of censoring boxes or application of significant blurring) to ensure appropriateness for submission.}
  \label{fig:supp_concepts}
  \vspace{-0.5cm}
\end{figure}

\begin{figure}[t]
\centering
  \includegraphics[width=0.9\linewidth]{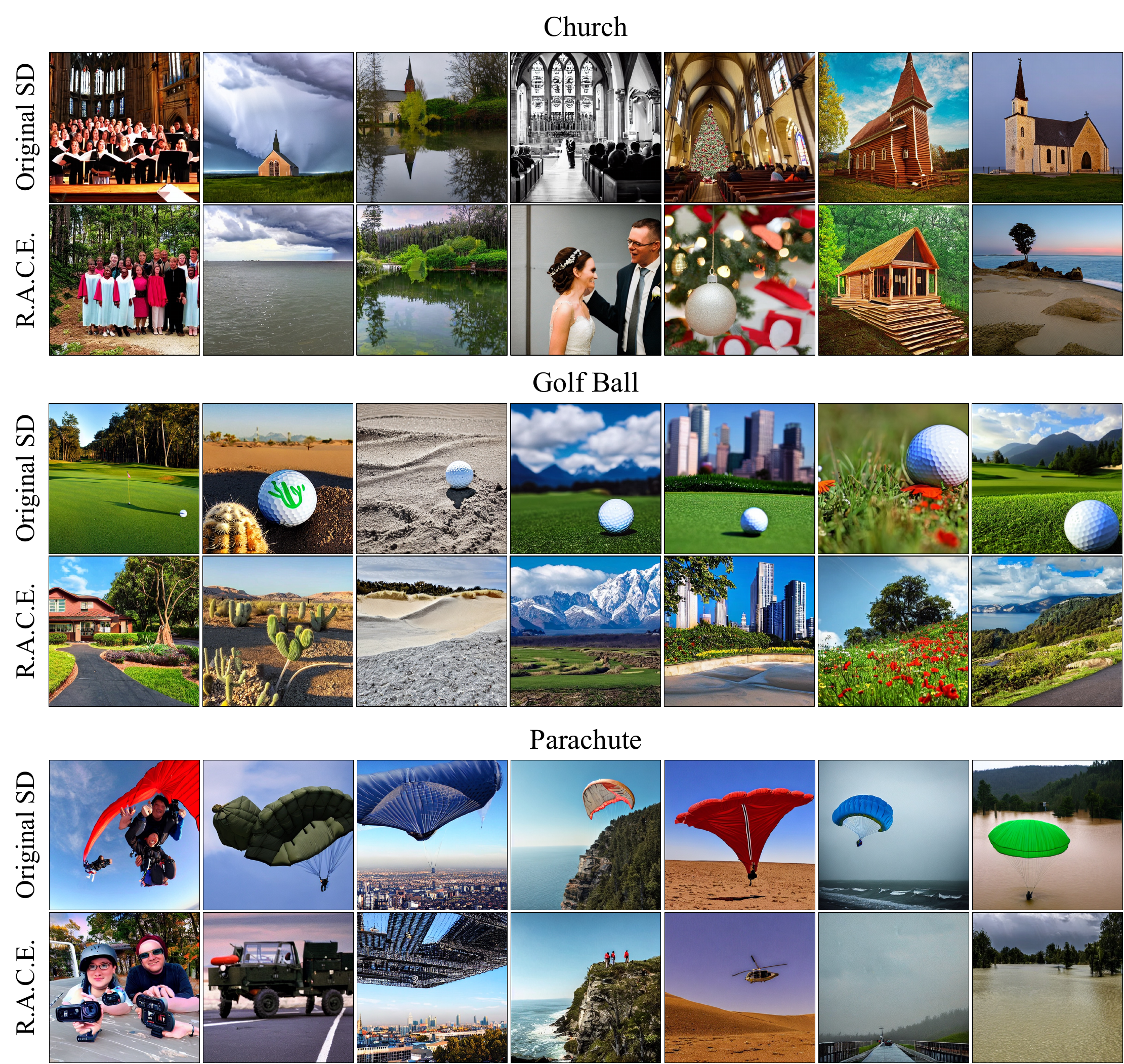}
  \caption{Comparison Between Original SD and RACE.}
  \label{fig:supp_objects}
  \vspace{-0.5cm}
\end{figure}

In Figures~\cref{fig:supp_gogh,fig:supp_concepts,fig:supp_objects}, we present a supplementary collection of images to further demonstrate the capability of our method in erasing targeted concepts. It is important to note that the set of images in~\cref{fig:supp_concepts} contains explicit content.

\begin{table}[t!]
\caption{Attack Success Rate (ASR) when varying $\epsilon$ and the number of adversarial attack steps. For each entry, we report three ASR values when $\epsilon=0.05, 0.1$, and $0.2$ using a tuple enclosed in parentheses.}
\centering
\begin{tabular}{L{2.8cm}C{4cm}C{4cm}}
\toprule
& \multicolumn{2}{c}{\# of adversarial attack steps} \\ \cmidrule(lr){2-3}
& 10 & 20  \\
\midrule
RACE-VanGogh & (0.88, 0.88, 0.88) & (0.90, 0.92, 0.94)  \\
RACE-Nudity & (0.94, 0.96, 0.99) & (0.94, 0.99, 0.97)  \\
RACE-Violence & (0.96, 0.98, 1.00) & (0.95, 0.99, 0.99) \\
RACE-Illegal & (0.92, 0.97, 1.00) & (0.95, 0.97, 0.97)  \\
RACE-Church & (0.70, 0.86, 0.90) & (0.74, 0.82, 0.96) \\
RACE-GolfBall & (0.74, 0.82, 0.96) & (0.32, 0.56, 0.74)\\
RACE-Parachute & (0.60, 0.68, 0.88) & (0.56, 0.76, 0.90) \\
\bottomrule
\end{tabular}
\label{tab:selection}
\end{table}

\section{Determining PGD Hyperparameters}
Within the main paper, we settle on $\epsilon = 0.1$ and designate 10 steps for the adversarial attack. This section delves into the empirical analysis underpinning this selection, particularly focusing on the balance between ASR efficacy and hyperparameter tuning. As delineated in~\cref{tab:selection}, an $\epsilon$ value of 0.2 marginally outperforms 0.1. However, as elaborated in the main manuscript, an elevated attack intensity with $\epsilon=0.2$ may compromise the equilibrium between adversarial robustness and the fidelity of generated images. Holding $\epsilon$ at 0.1, we observe a minimal variance between 10 and 20 attack steps, reinforcing our decision to configure PGD parameters at $\epsilon=0.1$ and 10 steps for an optimal trade-off.

\begin{wraptable}{r}{0.46\textwidth}
\vspace{-35pt}
\centering
\caption{Comparison of ASR decrease and CLIP-Score for different keywords}
\begin{adjustbox}{width=0.46\textwidth}
\begin{tabular}{lC{1.4cm}C{1.4cm}C{2cm}}
\toprule
Keyword & Nudity & Violence & Illegal Act \\
\midrule
ASR Decrease & -30\% & -11\% &-5\% \\ 
CLIP-Score~\cite{hessel2021clipscore} & 0.658 & 0.533 &0.526 \\
\bottomrule
\end{tabular}
\end{adjustbox}
\vspace{-20pt}
\label{tab:keyword-clipscore}
\end{wraptable}

\section{Challenges in Erasing Violence and Illegal Act}
The data presented in Table 2 of the main manuscript indicate that ``violence'' and ``illegal act'' concepts exhibit less pronounced reductions in Attack Success Rate (ASR) against UnlearnDiff white box attack~\cite{to_generate_or_not}. Specifically, ``nudity'' sees a 30\% decrease in ASR, while ``violence'' and ``illegal act'' experience reductions of 11\% and 5\%, respectively. To align with the experimental framework of UnlearnDiff~\cite{to_generate_or_not}, we employed ``violence'' and ``illegal act'' as keywords for concept removal. It is postulated that these terms may not optimally represent the range of hazardous imagery producible by Stable Diffusion (SD)~\cite{stable_diffusion} through the prompts in the I2P dataset~\cite{Schramowski2022SafeLD}. To examine this hypothesis, we calculate the CLIP-score~\cite{hessel2021clipscore} for explicit images generated by a baseline SD model, $SD(\Tilde{y})$, where $\Tilde{y}$ refers prompts from I2P and $c$ includes ``nudity'', ``violence'', and ``illegal act'' (i.e. CLIP-score($SD(\Tilde{y})$,$c$)). As illustrated in~\cref{tab:keywㅌord-clipscore}, a higher CLIP-score, indicating better keyword alignment with the dangerous images, correlates with a more substantial decrease in ASR. Conversely, keywords that poorly match the hazardous image content tend to result in less significant ASR reductions.
This underscores the critical importance of selecting highly representative keywords for concept erasure in T2I models, as their alignment with the generated content significantly influences the effectiveness of concept erase.

\newpage
%